%% file: main.tex
\documentclass[letterpaper, 10 pt, conference]{IEEE_styles/ieeeconf}  

\IEEEoverridecommandlockouts %only required if we need to use the \thanks command
\usepackage{0a_preamble}
\usepackage{pifont}

\begin{document}

%add the macros
\input{00_macros}

\input{0c_title}

\maketitle

%from Amir Template
\thispagestyle{empty}
\pagestyle{empty}

\input{0d_abstract}

\input{01_introduction}
\input{02_related_works}
\input{03_system_design}
\input{04_experiments}

\input{05_results}
\input{06_conclusion}

\bibliographystyle{IEEE_styles/IEEEtranMod}
\bibliography{references}

\end{document}

%% file: 00_macros.tex
%%%%%%%%%%%%%%%%%%%%%%%%%%%%%%%%%%%%%%%%%%%%%%%%%%%%%%%%%%%%%%%%%%%%%%%%%%%%
%%%%%%%%%%%%%%%%%%%%%%%%%%%%%%%%%%%%%%%%%%%%%%%%%%%%%%%%%%%%%%%%%%%%%%%%%%%%
% Project Descriptions/Names
%%%%%%%%%%%%%%%%%%%%%%%%%%%%%%%%%%%%%%%%%%%%%%%%%%%%%%%%%%%%%%%%%%%%%%%%%%%%
%%%%%%%%%%%%%%%%%%%%%%%%%%%%%%%%%%%%%%%%%%%%%%%%%%%%%%%%%%%%%%%%%%%%%%%%%%%%

\newcommand\ProjectName{RaGNNarok}

%%%%%%%%%%%%%%%%%%%%%%%%%%%%%%%%%%%%%%%%%%%%%%%%%%%%%%%%%%%%%%%%%%%%%%%%%%%%
%%%%%%%%%%%%%%%%%%%%%%%%%%%%%%%%%%%%%%%%%%%%%%%%%%%%%%%%%%%%%%%%%%%%%%%%%%%%
% Common Terms
%%%%%%%%%%%%%%%%%%%%%%%%%%%%%%%%%%%%%%%%%%%%%%%%%%%%%%%%%%%%%%%%%%%%%%%%%%%%
%%%%%%%%%%%%%%%%%%%%%%%%%%%%%%%%%%%%%%%%%%%%%%%%%%%%%%%%%%%%%%%%%%%%%%%%%%%%
%maps
\newcommand{\globalMap}{$m_{\textrm{G}}$}
\newcommand{\localMap}{$m_{\textrm{L}}$}
\newcommand{\localMapGF}{$\hat{m}_{\textrm{L}}$}

\newcommand\iwr{TI-IWR1843}
\newcommand\dca{TI-DCA1000}
\newcommand{\Lidar}{Lidar}
\newcommand{\lidar}{lidar}
\newcommand{\radar}{radar}
\newcommand{\Radar}{Radar}

%%%%%%%%%%%%%%%%%%%%%%%%%%%%%%%%%%%%%%%%%%%%%%%%%%%%%%%%%%%%%%%%%%%%%%%%%%%%
%%%%%%%%%%%%%%%%%%%%%%%%%%%%%%%%%%%%%%%%%%%%%%%%%%%%%%%%%%%%%%%%%%%%%%%%%%%%
% Localization/Filtering Terms
%%%%%%%%%%%%%%%%%%%%%%%%%%%%%%%%%%%%%%%%%%%%%%%%%%%%%%%%%%%%%%%%%%%%%%%%%%%%
%%%%%%%%%%%%%%%%%%%%%%%%%%%%%%%%%%%%%%%%%%%%%%%%%%%%%%%%%%%%%%%%%%%%%%%%%%%%
%kalman filter
\newcommand{\KalmanStateMatrix}{\textrm{\textbf{s}}}

%IMU Output
\newcommand{\accX}{a_{\textrm{x}}}
\newcommand{\accY}{a_{\textrm{y}}}
\newcommand{\accZ}{a_{\textrm{z}}}
\newcommand{\wPitch}{\omega_\Theta}
\newcommand{\wRoll}{\omega_\Phi}
\newcommand{\wYaw}{\omega_\Psi}

%IMU biases
\newcommand{\wYawBias}{\omega_{\textrm{bias}}}

%initial positions/headings
\newcommand{\poseInitial}{\textrm{\textbf{P}}_{\textrm{0}}}
\newcommand{\poseTransInitial}{\textrm{\textbf{p}}_{\textrm{0}}}
\newcommand{\poseHeadInitial}{\Psi_{\textrm{0}}}

%estimated positions/headings
\newcommand{\poseEst}[1]{\hat{\textrm{\textbf{P}}}_{#1}}
\newcommand{\poseTransEst}[1]{\hat{\textrm{\textbf{p}}}_{#1}}
\newcommand{\poseHeadEst}[1]{\hat{\Psi}_{#1}}

%ground truth positions/headings
\newcommand{\poseGT}[1]{{\textrm{\textbf{p}}_{#1}}}
\newcommand{\poseTransGT}[1]{\textrm{\textbf{p}}_{#1}}
\newcommand{\poseHeadGT}[1]{\Psi_{#1}}

%position
\newcommand{\pose}{\hat{\textrm{\textbf{p}}}}
\newcommand{\poseX}{x}
\newcommand{\poseY}{y}

%velocity
\newcommand{\vel}{\textrm{\textbf{v}}}
\newcommand{\velX}{v_{\textrm{x}}}
\newcommand{\velY}{v_{\textrm{y}}}

%orientation
\newcommand{\pitch}{\Theta}
\newcommand{\roll}{\Phi}
\newcommand{\yaw}{\Psi}

%ugv buffer radius
\newcommand{\bufferUGV}{\textrm{R}_{\textrm{UGV}}}

%%%%%%%%%%%%%%%%%%%%%%%%%%%%%%%%%%%%%%%%%%%%%%%%%%%%%%%%%%%%%%%%%%%%%%%%%%%%
%%%%%%%%%%%%%%%%%%%%%%%%%%%%%%%%%%%%%%%%%%%%%%%%%%%%%%%%%%%%%%%%%%%%%%%%%%%%
% Paragraph Labels
%%%%%%%%%%%%%%%%%%%%%%%%%%%%%%%%%%%%%%%%%%%%%%%%%%%%%%%%%%%%%%%%%%%%%%%%%%%%
%%%%%%%%%%%%%%%%%%%%%%%%%%%%%%%%%%%%%%%%%%%%%%%%%%%%%%%%%%%%%%%%%%%%%%%%%%%%
\newcommand{\paragraphLabel}[1]{\noindent\textbf{#1}}
\newcommand{\emphasizedLabel}[1]{\noindent\textit{#1}}
\newcommand{\keyStep}[2]{\noindent\textbf{Step \circled{#1}: #2}}
\newcommand{\softwarePackage}[1]{\texttt{#1}}

%%%%%%%%%%%%%%%%%%%%%%%%%%%%%%%%%%%%%%%%%%%%%%%%%%%%%%%%%%%%%%%%%%%%%%%%%%%%
%%%%%%%%%%%%%%%%%%%%%%%%%%%%%%%%%%%%%%%%%%%%%%%%%%%%%%%%%%%%%%%%%%%%%%%%%%%%
% figure, section, eq, and table Labels
%%%%%%%%%%%%%%%%%%%%%%%%%%%%%%%%%%%%%%%%%%%%%%%%%%%%%%%%%%%%%%%%%%%%%%%%%%%%
%%%%%%%%%%%%%%%%%%%%%%%%%%%%%%%%%%%%%%%%%%%%%%%%%%%%%%%%%%%%%%%%%%%%%%%%%%%%
\newcommand{\sect}[1]{Sec.~{#1}}
\newcommand{\fig}[1]{Fig.~{#1}}
\newcommand{\eq}[1]{eq.~{#1}}
\newcommand{\tbl}[1]{Table~{#1}}

%%%%%%%%%%%%%%%%%%%%%%%%%%%%%%%%%%%%%%%%%%%%%%%%%%%%%%%%%%%%%%%%%%%%%%%%%%%%
%%%%%%%%%%%%%%%%%%%%%%%%%%%%%%%%%%%%%%%%%%%%%%%%%%%%%%%%%%%%%%%%%%%%%%%%%%%%
%Units
%%%%%%%%%%%%%%%%%%%%%%%%%%%%%%%%%%%%%%%%%%%%%%%%%%%%%%%%%%%%%%%%%%%%%%%%%%%%
%%%%%%%%%%%%%%%%%%%%%%%%%%%%%%%%%%%%%%%%%%%%%%%%%%%%%%%%%%%%%%%%%%%%%%%%%%%%
\newcommand{\dBsm}[1]{{#1}\thinspace{dBsm}}
\newcommand{\dBsmSq}[1]{{#1}\thinspace{$\textrm{dBsm}^2$}}
\newcommand{\dB}[1]{{#1}\thinspace{dB}}
\newcommand{\dBm}[1]{{#1}\thinspace{dBm}}
%times
\newcommand{\nsec}[1]{{#1}\thinspace{ns}}
\newcommand{\usec}[1]{{#1}\thinspace{$\upmu$s}}
\newcommand{\msec}[1]{{#1}\thinspace{ms}}
\newcommand{\seconds}[1]{{#1}\thinspace{s}}
%frequencies
\newcommand{\GHz}[1]{{#1}\thinspace{GHz}}
\newcommand{\MHz}[1]{{#1}\thinspace{MHz}}
\newcommand{\kHz}[1]{{#1}\thinspace{kHz}}
\newcommand{\Hz}[1]{{#1}\thinspace{Hz}}
%Data Rates
\newcommand{\MSps}[1]{{#1}\thinspace{MSps}}
\newcommand{\MBps}[1]{{#1}\thinspace{MBps}}

%slopes
\newcommand{\MHzPerus}[1]{{#1}\thinspace{MHz/$\upmu$s}}
%distances
\newcommand{\m}[1]{{#1}\thinspace{m}}
\newcommand{\mSq}[1]{{#1}\thinspace{$\textrm{m}^2$}}
\newcommand{\cm}[1]{{#1}\thinspace{cm}}
\newcommand{\mm}[1]{{#1}\thinspace{mm}}
%velocities
\newcommand{\mPers}[1]{{#1}\thinspace{m/s}}

%angles
\newcommand{\degrees}[1]{{#1}$^{\circ}$}
\newcommand{\radians}[1]{{#1}\thinspace{rad}}
\newcommand{\radiansSq}[1]{${#1\textrm{rad}}^{2}$}

%mass
\newcommand{\grams}[1]{{#1}\thinspace{g}}
\newcommand{\kilograms}[1]{{#1}\thinspace{kg}}

%power
\newcommand{\Watts}[1]{{#1}\thinspace{W}}

%constants
\newcommand{\LightSpeed}{\textrm{c}}

%%%%%%%%%%%%%%%%%%%%%%%%%%%%%%%%%%%%%%%%%%%%%%%%%%%%%%%%%%%%%%%%%%%%%%%%%%%%
%%%%%%%%%%%%%%%%%%%%%%%%%%%%%%%%%%%%%%%%%%%%%%%%%%%%%%%%%%%%%%%%%%%%%%%%%%%%
% Radar Parameters,Performance Specificeations, detections
%%%%%%%%%%%%%%%%%%%%%%%%%%%%%%%%%%%%%%%%%%%%%%%%%%%%%%%%%%%%%%%%%%%%%%%%%%%%
%%%%%%%%%%%%%%%%%%%%%%%%%%%%%%%%%%%%%%%%%%%%%%%%%%%%%%%%%%%%%%%%%%%%%%%%%%%%
%detections
\newcommand{\detection}{\textrm{\textbf{d}}}
\newcommand{\pointCloud}[1]{\mathcal{C}_{\textrm{#1}}}

\newcommand{\detectionX}{d_{\textrm{x}}}
\newcommand{\detectionY}{d_{\textrm{y}}}
\newcommand{\detectionZ}{d_{\textrm{z}}}
\newcommand{\detectionV}{d_{\textrm{v}}}
\newcommand{\detectionP}{p_{\textrm{det}}}

%GNN nodes
\newcommand{\node}{\textrm{\textbf{n}}}

%% Radar Signals
\newcommand{\TxSignal}{s_{\textrm{TX}}(t)}
\newcommand{\RxSignal}{s_{\textrm{RX}}(t)}
\newcommand{\RxSignalConj}{s_{\textrm{RX}}^{*}(t)}
\newcommand{\IFSignal}{s_{\textrm{IF}}(t)}

%% Radar Configuration
\newcommand{\NumChirpsPerFrame}{N_{\textrm{chirps}}}
\newcommand{\FrameDuration}{T_{\textrm{frame}}}
\newcommand{\ChirpDuration}{T_{\textrm{chirp}}}
\newcommand{\ChirpFreqStart}{f_{c}}
\newcommand{\ChirpWavelength}{\lambda}
\newcommand{\ChirpSlope}{S}
\newcommand{\ChirpSlopeVic}{S_{\textrm{vic}}}
\newcommand{\ChirpSlopeAtt}{S_{\textrm{atk}}}
\newcommand{\ChirpBW}{B}
\newcommand{\FreqSamp}{f_{\textrm{samp}}}
\newcommand{\NSamps}{N_{\textrm{Samp}}}
\newcommand{\NChirps}{N_{\textrm{Chirps}}}
\newcommand{\KRx}{K_{\textrm{Rx}}}
\newcommand{\KTx}{K_{\textrm{Tx}}}

% Performance Specifications
\newcommand{\RangeRes}{d_{\textrm{res}}}
\newcommand{\RangeMax}{d_{\textrm{max}}}
\newcommand{\RangeMin}{d_{\textrm{min}}}
\newcommand{\VelocityRes}{v_{\textrm{res}}}
\newcommand{\VelocityMax}{v_{\textrm{max}}}
\newcommand{\AngleRes}{\theta_{\textrm{res}}}

%radar target
\newcommand{\TargetRange}{d_{\textrm{obj}}}
\newcommand{\TargetVelocity}{v_{\textrm{obj}}}
\newcommand{\TargetAngle}{\theta_{\textrm{obj}}}

% Intermediate terms (time delays, phase shifts, IF frequencies)
\newcommand{\tDelay}{t_{\textrm{d}}}
\newcommand{\TargetPhaseShift}{\phi_{\textrm{obj}}}
\newcommand{\DopplerShift}{\phi_{\textrm{doppler}}}
\newcommand{\FreqIF}{f_{\textrm{IF}}}
%%%%%%%%%%%%%%%%%%%%%%%%%%%%%%%%%%%%%%%%%%%%%%%%%%%%%%%%%%%%%%%%%%%%%%%%%%%%
%%%%%%%%%%%%%%%%%%%%%%%%%%%%%%%%%%%%%%%%%%%%%%%%%%%%%%%%%%%%%%%%%%%%%%%%%%%%
% RadNav Parameters
%%%%%%%%%%%%%%%%%%%%%%%%%%%%%%%%%%%%%%%%%%%%%%%%%%%%%%%%%%%%%%%%%%%%%%%%%%%%
%%%%%%%%%%%%%%%%%%%%%%%%%%%%%%%%%%%%%%%%%%%%%%%%%%%%%%%%%%%%%%%%%%%%%%%%%%%%

%% Test sites
\newcommand{\Wilkinson}{\ding{182}}
\newcommand{\NorthVICON}{\ding{183}}
\newcommand{\NorthBasement}{\ding{184}}
\newcommand{\CPSL}{\ding{185}}

%% Ground vehicle parameters
\newcommand{\UGVRadius}{R_{UGV}}
%%%%%%%%%%%%%%%%%%%%%%%%%%%%%%%%%%%%%%%%%%%%%%%%%%%%%%%%%%%%%%%%%%%%%%%%%%%%
%%%%%%%%%%%%%%%%%%%%%%%%%%%%%%%%%%%%%%%%%%%%%%%%%%%%%%%%%%%%%%%%%%%%%%%%%%%%
% Comments
%%%%%%%%%%%%%%%%%%%%%%%%%%%%%%%%%%%%%%%%%%%%%%%%%%%%%%%%%%%%%%%%%%%%%%%%%%%%
%%%%%%%%%%%%%%%%%%%%%%%%%%%%%%%%%%%%%%%%%%%%%%%%%%%%%%%%%%%%%%%%%%%%%%%%%%%%%create flags to show comments/todos
\newif\ifShowComments
\newif\ifShowToDos

%set flags
\ShowCommentstrue
\ShowToDostrue

%david comments
\newcommand\david[1]{
    \ifShowComments\textcolor{blue}{[DH: #1]} \fi%
}

%shaocheng comments
\newcommand\shaocheng[1]{
    \ifShowComments\textcolor{purple}{[SL: #1]} \fi%
}

%% file: 0c_title.tex
\title{\LARGE \bf 
\ProjectName: A Light-Weight Graph Neural Network for Enhancing Radar Point Clouds on Unmanned Ground Vehicles}

%create flags to show comments/todos
\newif\ifAnonymize

%set flags
\Anonymizefalse

%add title if Anonymize is false
\ifAnonymize

\else
    \author{David Hunt*, Shaocheng Luo*, Spencer Hallyburton, Shafii Nillongo, Yi Li, \\ Tingjun Chen, and Miroslav Pajic  %
    \thanks{*These authors contributed equally to this work.}
    \thanks{This work is sponsored in part by the ONR under the agreement N00014-23-1-2206, AFOSR FA9550-19-1-0169 Award, NSF CNS-1652544 and CNS-2211944 awards, and the National AI Institute for Edge Computing Leveraging Next Generation Wireless Networks, Grant CNS-2112562.}% <-this % stops a space
    \thanks{The authors are with the Department of Electrical and Computer
    Engineering, Duke University, Durham, NC 27708 USA (e-mail:
    \{david.hunt, shaocheng.luo, spencer.hallyburton, shafii.nillongo, yi.li, tingjun.chen, miroslav.pajic\}@duke.edu).}
    }
\fi

%% file: 0d_abstract.tex
%%%
% SECTION IS NOW LOCKED - MIROSLAV HAS ALREADY REVIEWED THIS SECTION SO PLEASE DON'T MAKE FURTHER CHANGES BEYOND TYPOS
%%%

%
% I see the note above, but logically, it is a tautology to basically say "UGVs are used where UGVs are needed" in the first line, and tautologies are bad logic, so I recommend a slight edit.
%

\begin{abstract}
% Unmanned ground vehicles (UGVs) are widely used in applications where a UGV must determine its location and navigate in a known environment. Currently, camera and light detection and ranging ({\lidar}) sensors are used, but they have limitations such as poor performance in visually obscured environments, high computational overhead for data~processing, and high costs for {\lidar}s. Conversely, mmWave {\radar} sensors are affordable, lightweight, and provide accurate ranging regardless of visibility. While existing methods focused on {\radar} \emph{odometry} (i.e., determining relative position over time), we introduce {\ProjectName}, a {\radar}-based \emph{navigation} framework that enables \emph{low-cost} \emph{resource-constrained} UGVs to localize and navigate through complex and dynamic indoor environments. Using our novel point cloud processing approach, we show how a UGV can accurately localize itself in known environments, outperforming traditional and state-of-the-art learning-based methods. Also, {\ProjectName} consistently plans collision-free trajectories that avoid unmapped static and dynamic obstacles using a hierarchical path planner. Finally, we demonstrate {\ProjectName}'s effectiveness on a real-time prototype operating in a range of complex indoor environments. To~the~best~of~our~knowledge, \ProjectName~is the first mmWave {\radar}-based indoor navigation framework.

Low-cost indoor mobile robots have gained popularity with the increasing adoption of automation in homes and commercial spaces. However, existing {\lidar} and camera-based solutions have limitations such as poor performance in visually obscured environments, high computational overhead for data~processing, and high costs for {\lidar}s. In contrast, mmWave {\radar} sensors offer a cost-effective and lightweight alternative, providing accurate ranging regardless of visibility. However, existing radar-based localization suffers from sparse point cloud generation, noise, and false detections. Thus, in this work, we introduce {\ProjectName}, a real-time, lightweight, and generalizable graph neural network (GNN)-based framework to enhance radar point clouds, even in complex and dynamic environments. With an inference time of just 7.3 ms on the low-cost Raspberry Pi~5, {\ProjectName} runs efficiently even on such resource-constrained devices, requiring no additional computational resources. We evaluate its performance across key tasks, including localization, SLAM, and autonomous navigation, in three different environments. Our results demonstrate strong reliability and generalizability, making {\ProjectName} a robust solution for low-cost indoor mobile robots.

\end{abstract}

%% file: 01_introduction.tex
\section{Introduction}
\label{sec:introduction}

%%%%%%%%%%%%%%%%%%%%%%%%%%%%%%%%%
% PARAGRAPH 1
%%%%%%%%%%%%%%%%%%%%%%%%%%%%%%%%%

Indoor mobile robots, such as unmanned ground vehicles (UGVs), are increasingly used in homes and commercial spaces, requiring accurate sensing for GNSS-free mapping and navigation in potentially complex environments. Traditionally, these robots rely on {\lidar} and cameras, but both technologies have limitations that hinder their wide adoption on low-cost platforms.

% {\lidar} 
Lidar offers high precision but is expensive (e.g., \$800+ for the Livox MID-360), power-intensive, and struggles in visually occluded environments (e.g., smoke, dust). Cameras, while affordable, face challenges in low-light or visually uniform environments and lack depth perception unless combined with additional sensors. Both {\lidar} and camera-based systems require deep learning (DL) models for reliable scene understanding, which may exceed the computational capacity of low-cost robots and fail in tasks that have real-time~requirements.

In contrast, mmWave radar provides a low-cost, lightweight alternative with accurate ranging, even in poor visibility. Modern 77 GHz mmWave radars achieve 4 cm range resolution in a compact, low-power form factor, making them ideal for indoor mobile robots. Unlike {\lidar} and cameras, radar can also detect velocity, enabling real-time differentiation between static and dynamic objects, and enhancing situational awareness in complex environments.

While mmWave radar may offer a low-cost alternative for UGV-based mapping, localization, and navigation, several critical challenges have hindered its real-time adoption. First, the angular resolution of typical mmWave radar sensors used on UGVs is limited to 14.3° \cite{rao_introduction_nodate}, significantly coarser than {\lidar}'s 0.1° resolution \cite{velodyne_lidar_vlp-16_2018}. This results in 90\% fewer points than even 2D {\lidar} slices, leading to sparse and incomplete environmental representations. Second, radar point clouds are prone to high false detection rates due to multipath interference, where radio waves reflect off multiple objects before returning to the sensor. In indoor environments, we observed that up to \underline{\emph{60\% of detected points were false}}, further complicating localization and mapping.

Existing approaches for mmWave radar-based sensing on UGVs struggle with real-time feasibility and degrade in dynamic environments. Some methods encode radar inputs as images\cite{prabhakara_high_2023,Sie_radarize_2024,hunt_2024_RadCloud}, while others rely on generative adversarial networks (GANs)~\cite{lu_see_2020}, but these approaches demand high-compute, segmentation-based deep learning models that are impractical for UGVs with limited computational resources. Additionally, no prior work effectively utilizes radar velocity measurements, making it challenging to distinguish static from moving objects—a crucial limitation that hampers mmWave radar-only  mapping %building
and reliable navigation.

To overcome these challenges, we introduce {\ProjectName}, a lightweight graph neural network (GNN)-based framework that enhances 2D mmWave radar point clouds for real-time UGV navigation. Unlike CNN-based approaches that struggle with irregular, sparse radar data, {\ProjectName} directly models point relationships as a graph, enabling spatial feature aggregation while filtering multipath artifacts. By incorporating velocity measurements, our method uniquely discriminates between static and dynamic objects, improving navigation robustness. Additionally, its efficient architecture significantly boosts frame rate, achieving 7.3 ms inference time on a Raspberry Pi 5, making {\ProjectName} the first practical GNN-based solution for real-time mmWave radar-based UGV autonomy.

To demonstrate both performance and real-time feasibility, we integrate {\ProjectName} into the navigation pipeline on a resource-constrained UGV. Combined with industry---standard ROS2 packages—Nav2 and slam-toolbox (\cite{macenski_nav2_2023survey,macenski_nav2_2020marathon2,Macenski_slam_toolbox_2021})---we show that our GNN-enhanced mmWave radar framework enables accurate mapping, real-time localization, and robust navigation in complex, dynamic environments. Unlike prior methods that struggle with sparse, noisy radar data, {\ProjectName} enhances point clouds in real-time, ensuring precise localization without reliance on high-resolution {\lidar} or vision-based sensors.
To the best of our knowledge, this is the first work to achieve real-time localization and navigation on a UGV using enhanced mmWave radar point clouds, demonstrating the practical viability of GNNs for mmWave radar-based~autonomy.

This paper is organized as follows. Section \sect{\ref{sec:system_design}} introduces the framework used to implement the {\ProjectName} model. Following, section \sect{\ref{sec:experiments}} describes the robust evaluations used to validate {\ProjectName}, including real-world case studies on a UGV and comparisons to existing traditional and learning-based methods. Finally, section \sect{\ref{sec:results}} presents the results of our evaluations where we demonstrate the accuracy, computational efficiency, and feasibility of the {\ProjectName} model. 

%% file: 02_related_works.tex
\section{Related Works}
\label{sec:related_works}

\paragraphLabel{Deep learning (DL) models for mmWave radar sensing.}
Previous works~ \cite{prabhakara_exploring_2022,hunt_2024_RadCloud,Sie_radarize_2024,lu_see_2020,cai_millipcd_2022,geng_dream_2024} have introduced methods of converting low resolution mmWave {\radar} data into high-resolution 2D and/or 3D \lidar-like point clouds through various DL models. However, these models only focused on static environments and either do not generalize well in complex, previously unseen, environments and/or cannot be executed in real-time on computationally-constrained platforms. 
Furthermore, \cite{Lai_PanoRadar_2024} generated high-resolution 3D reconstructions on indoor environments by using a rotation mmWave {\radar}, but also only focused on static environments and requires significant computational resources to run in real time. By contrast, {\ProjectName} allows for operating in complex and dynamic environments while being generalizable and executing in real-time on resource-constrained platforms. 

\paragraphLabel{Graph Neural Networks (GNNs).}
Previously, \cite{svenningsson_radar_2021, sevimli_graph_2023, Fent_RadarGNN_2023
} have each applied GNNs to mmWave {\radar} point clouds; yet, they focused only on detecting specific objects (e.g., vehicles and people) in {\radar} point clouds. By contrast, {\ProjectName} is designed to enhance the point cloud corresponding \emph{only} to the static environment by identifying and filtering out multi-path detections. 

\paragraphLabel{Automotive \radar s.} In the automotive domain, previous works have implemented mmWave{\radar}-based localization \cite{adams_robotic_2012,Narula_2020_automotive,Abosekeen_2018_utilizing,Ort_2020_Autonomous,cen_2018_Precise,Pishehvari_2019_radar,Rouveure_2019_robot,Hong_2022_radarslam,Adolfsson_2022_lidar}, using methods like the cross-correlation of occupancy grid maps~\cite{Narula_2020_automotive} and the Fourier-Mellin Transform method~\cite{Pishehvari_2019_radar}. Further, ~\cite{Rouveure_2019_robot,Hong_2022_radarslam,adams_robotic_2012} introduced simultaneous localization and mapping (SLAM) pipelines. Yet, all focused on outdoor environments and used \emph{high-resolution} {\radar}s such as the CTS350-X~\cite{navtech_CTS350_2023} with angular resolutions ranging from \degrees{1} to \degrees{4}. These {\radar}s are expensive, large, relatively heavy (\kilograms{6}), consume a high amount of power (\Watts{24}), and require high data rates (1\thinspace{Gbps} ethernet), making them infeasible for low-cost, resource-constrained UGVs~\cite{navtech_CTS350_2023}.

\vspace{2pt}
\paragraphLabel{\Radar\ odometry.}
Regarding UGV odometry and collision avoidance, \cite{chen_2023_DRIO,Lu_2020_milliEgo,almalioglu_2020_milliRio,Lu_2022_lowcost,Doe_2021_xrio,Doer_2020_ekf,Park_2021_3d} proposed techniques for mmWave {\radar}-based \emph{odometry} that determines a UGV's relative location with regards to its starting point in an unknown~environment; e.g.,~\cite{chen_2023_DRIO,Lu_2022_lowcost,Doe_2021_xrio,Doer_2020_ekf,Park_2021_3d} used a mmWave {\radar} to estimate the velocity and an IMU sensor to determine the yaw angle and rate of the vehicle. Still, these methods are susceptible to large odometry drifts over time making them infeasible for localization over longer trajectories.

\vspace{2pt}
\paragraphLabel{Collision avoidance.} Recently, \cite{Reich_2020_memory,schouten2019biomimetic}
developed mmWave radar-based collision avoidance for indoor UGVs. Yet, both were restricted to static environments; \cite{Reich_2020_memory} only operates in simple small-scale rectangular spaces, and \cite{schouten2019biomimetic} only plans~continuous paths around an environment that avoids~collisions. 

%% file: 03_system_design.tex
\section{System Design}
\label{sec:system_design}

{\ProjectName} utilizes a UGV's estimated position and velocity (e.g., obtained from wheel encoders) to efficiently enhance noisy {\radar} point clouds and differentiate between static and dynamic detections (see \fig{\ref{fig:block_diagram}}). The enhanced point clouds can then be used for downstream tracking, mapping, localization, and navigation tasks in real-time. We now introduce the three main {\ProjectName} components. 
\input{figures/tex_files/block_diagram}

\subsection{{\Radar} Point Cloud Pre-processing}

To optimize performance of the {\ProjectName} framework, we perform the following pre-processing steps. 

\keyStep{1}{Multi-\radar\ point cloud.}
First, we utilize a ``front" and ``rear" \GHz{77} {\iwr} {\radar} sensor \cite{texas_instruments_iwr1843_2018,texas_instruments_iwr1843_2020} operating at \Hz{20} to achieve a \degrees{360} field-of-view (FOV). This provides greater situational awareness and significantly improves down-stream mapping, localization, and navigation performance when one {\radar} is obscured (e.g., by objects or people close to the vehicle). Here, each {\radar} detection ($\detection = [\detectionX; \detectionY; \detectionZ; \detectionV]$) contains the 3D Cartesian coordinates and relative velocity (i.e., velocity towards or away from the \radar) of an object. 

While the {\iwr} can perform 3D sensing, we opt to use a 2D configuration (i.e., $\detectionZ$ term is set to 0) as we found the elevation estimates to be inaccurate for UGV scenarios. Thus, both {\radar}s are configured to have a range resolution ($\RangeRes$) of~\cm{7}, maximum sensing range ($\RangeMax$) of \m{8.56}, azimuth angular resolution ($\AngleRes$) of \degrees{14.3}, and velocity resolutions ($\VelocityRes$) \mPers{0.01}. Here, we minimize the processing load by using the TI mmWave SDK \cite{texas_instruments_mmwaveSDK_2024} to implement a standard {\radar} processing pipeline directly on the {\radar}s. Additionally, we filter ground detections by removing all detections within a range of \m{1.5}. The final combined point cloud typically features $\sim${50} points.

\keyStep{2}{Dynamic object detection.}
In dynamic environments, a significant number of {\radar} detections can correspond to moving objects (e.g., people). Thus, for each new frame of {\radar} data, we identify detections corresponding to dynamic objects by utilizing the UGV's translational velocity ($\vel$) and the velocity component of each detection ($\detectionV$). For a detection with 2D coordinates $\textbf{q}=[\detectionX; \detectionY]$, the relative velocity measured by the \radar\ for a static point can be expressed as
$
%\hat{\detectionV} = -1 \cdot \frac{\vec{q}}{||\vec{q}||_{2}} \cdot \vel.
\hat{\detectionV}=\langle -\frac{\textbf{q}}{\|\textbf{q}\|}, \vel \rangle
$
Thus, we define dynamic detections as any detection where $|\hat{\detectionV} - \detectionV| > 0.05 \, \text{m/s}$. Once separated, static detections are then used to generate a probabilistic occupancy grid map while dynamic detections can then be used by other downstream tasks (e.g.; tracking) 
%NOTE: we didn't implement tracking, but we do publish the dynamic detections

\keyStep{3}{Probabilistic occupancy grid}
Given the sparsity of the {\radar} point clouds, we additionally employ a probabilistic occupancy grid to provide recent temporal history to the model. Empirically, we found that an occupancy grid with a cell resolution of \cm{20}, range of [\m{-5},\m{5}], and temporal history leveraging the previous 20 {\radar} frames (i.e. \seconds{1} of previous sensing) best balanced between increasing point cloud accuracy while minimizing the required sensing duration. Here, we utilize a UGV's pose (i.e.; position and orientation) estimate to continuously align the occupancy grid with the most recently recorded {\radar} frame. 

\subsection{{\ProjectName} Model}
\input{figures/tex_files/gnn_overview}

Graph neural networks (GNNs) encode data as a set of nodes and edges where each node is defined by a set of properties and each edge defines how the nodes are connected. Compared to other methods, graph neural networks are particularly well suited for enhancing {\radar} point clouds because their graph structure allows them to work well with sparse data and spatial representations while also being more robust to noise and false detections \cite{Zhou_graph_2020}. {\ProjectName} takes in a {\radar} point cloud and detection probability information to classify each detection as valid or invalid (e.g.; a multipath detection).

\vspace{4pt}
\paragraphLabel{Input graph nodes and edges} 
For each {\radar} detection captured by the probabilistic occupancy grid, we define a  node ($\node = [\detectionX; \detectionY; \detectionZ; \detectionP]$) which contains the Cartesian coordinates and current probability of the detection. For the edges between nodes, we use the Pytorch Geometric \softwarePackage{radius\_graph} module to define the edges between all nodes within a \m{10} radius of each other. Here, each edge's value is the euclidean distance between the two corresponding nodes. Compared to previous segmentation approaches, we highlight how this input format significantly reduces the size of the model input data while allowing it to dynamically adapt to point clouds with varying numbers of points. 

\paragraphLabel{Model Output.}
For each {\radar} node in the graph, {\ProjectName} classifies each node as valid or invalid. When labeling each node, we define a node as valid if its corresponding {\radar} detection was within \cm{20} of a {\lidar} ground-truth detection.

\paragraphLabel{Model Architecture.} 
As shown in \fig{\ref{fig:gnn_overview}}, we implemented a light-weight GNN architecture by using a series of three GraphSAGE convolution blocks using the Pytorch \softwarePackage{SAGEConv} modules \cite{hamilton_sagegnn_2017}. Compared to other graphical convolution methods, we selected GraphSAGE convolutions because they learn a flexible aggregation function and generalize particularly well beyond the training data \cite{hamilton_sagegnn_2017}. In total, our model only utilizes 705 parameters, compared to recent state of the art (SOTA) works \cite{hunt_2024_RadCloud} and \cite{prabhakara_high_2023} which utilized $\sim${7.7}\thinspace{M} and $\sim${17.5}\thinspace{M} parameters, respectively. As we show in the following sections, this simpler model significantly reduces the computational time required for each inference. 

\paragraphLabel{Loss Function}
We utilize Binary Cross Entropy (BCE) loss during training as this is a commonly applied loss function in graph neural network training pipelines and is particularly well suited for classification tasks. 

\paragraphLabel{Data Augmentations}
To further improve the robustness of our model as well as to prevent over-fitting during training, we additionally employed the following three data augmentations during the training process. First, a random yaw (i.e., around the z axis) rotation in the range [\degrees{0},\degrees{360}]. Secondly, we applied a random perturbations to the occupancy probability (i.e., $\detectionP$) of each node by sampling from a normal distribution with a standard deviation of 0.05. Finally, we randomly perturbed  $(\detectionX, \detectionY)$ for each node by independently sampling from a normal distribution with standard deviation of \cm{16}. 

\subsection{Detection History}
After enhancing the {\radar} point clouds using the {\ProjectName} model, we further increase the density of the final output point cloud by maintaining a short history of detected points. To achieve this, we keep a list of valid detections, and each detection is maintained for a duration of 10 {\radar} frames (i.e., \seconds{0.5} worth of {\radar} frames).  As with the probabilistic occupancy grid, we utilize the UGVs pose information to continuously align the point cloud history with the most recently received {\radar} frame. 

%% file: figures/tex_files/block_diagram.tex
% column wide figure
\begin{figure}[t!]
\centering
\includegraphics[width=0.98\columnwidth]{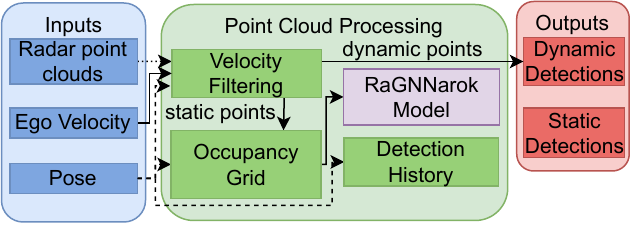}
% \vspace{-4pt}
\caption{\ProjectName~block diagram.}
\label{fig:block_diagram}
\end{figure}

%% file: figures/tex_files/gnn_overview.tex
\begin{figure*}[!t]
\centering
\includegraphics[width=1.94\columnwidth]{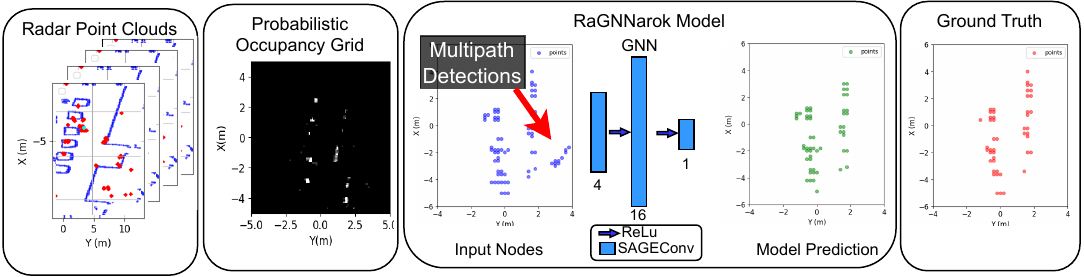} 
\caption{Overview of the {\ProjectName} model architecture}
\label{fig:gnn_overview}
\end{figure*}

%% file: 04_experiments.tex
%%%
% SECTION IS NOW LOCKED - MIROSLAV HAS ALREADY REVIEWED THIS SECTION SO PLEASE DON'T MAKE FURTHER CHANGES BEYOND TYPOS
%%%

\section{Evaluation}
\label{sec:experiments}
We rigorously validated the performance of {\ProjectName} using a real-world ROS2 prototype UGV equipped compute-limited single board computer. In addition to assessing the quality of the generated point clouds, we also evaluated the performance of downstream localization, navigation, and mapping tasks through a series of offline and real-time experiments. Furthermore, we benchmark our performance against traditional and state of the art DL models. Notably, evaluations were performed across a diverse set of complex environments where people were regularly moving throughout the scene.

\subsection{Experimental Setup}

\paragraphLabel{Platform.}
We used an iRobot Create3 unmanned ground vehicle (\fig{\ref{fig:platform_overview}}) equipped with a Raspberry Pi5 single board computer, \iwr front and back {\radar}s, a Livox MID-360 {\lidar} (used only for collecting ground truth), and Xsens MTi-10 IMU sensor. Radar and \lidar\ data were sampled at a rate of \Hz{20} while IMU measurements were sampled at \Hz{200}. Also, we align the coordinate frames of the sensors by applying coordinate transformations to the {\radar} data such that both {\radar}s have the same coordinate frame as the {\lidar}. The entire design is implemented in a real-time~ROS2~framework.

\input{figures/tex_files/platform_overview}

\vspace{2pt}
\paragraphLabel{Test environments.}
\input{figures/tex_files/test_environments}
We used the three complex environments shown in \fig{\ref{fig:test_environments}} to evaluate the performance of \ProjectName. Each environment featured people moving freely throughout the scene, objects of different shapes, sizes, and materials, and a combination of enclosed and open spaces. Finally, we also ensured that \ProjectName\ is robust to changing environments by moving various objects (e.g., chairs, tables, and whiteboards) in all environments between trials.

\subsection{Baseline Methods}
{\ProjectName}'s performance is compared against the following baseline methods.

\vspace{2pt}
\paragraphLabel{Naive \Radar}: Utilizing the combined point cloud from the \radar\ sensors (i.e., no additional filtering or point cloud stacking) to localize the UGV at each frame.

\vspace{2pt}
\paragraphLabel{RadarHD~\cite{prabhakara_high_2023}:} A DL segmentation model that converts~raw data from a single \radar\ into 2D \lidar-like point clouds. Here, we utilized a \dca\ to capture raw data from the front \radar\ and RadarHD's pre-trained model to obtain a predicted 2D point cloud to localize the UGV at each frame.

\vspace{2pt}
\paragraphLabel{RadCloud~\cite{hunt_2024_RadCloud}:} A more recent DL model, %specifically 
optimized for resource-constrained platforms, that converts raw data from a single \radar\ into 2D \lidar-like point clouds. %As with RadarHD, we utilized 
Again, we used the \dca\ to capture raw data from the front \radar\ and RadCloud's pre-trained model to obtain a predicted 2D point cloud for localizing the UGV in each frame.

\subsection{Datasets}
For model training, point cloud quality evaluation, and localization accuracy experiments, we recorded the following large-scale datasets. 

\paragraphLabel{{\ProjectName} Training, Validation, and Test Datasets.}
For training, validation, and testing of the {\ProjectName} model, we captured a total of 57,641 time-synchronized frames by performing 33 independent trials across the three testing environments, covering a total trajectory distance of 1.3 km. For each trial, the UGV was driven along a unique trajectory consisting of irregular turns at varying speeds. All trials contained people moving freely throughout the scene and objects that changed position from trial to trial. For model training and validation, we used 5,400 samples for training and 1,801 samples for validation. Notably, only environment 2 was used for training the model, allowing environments 1 and 3 to be used for assessing performance in new environments. The remaining 50, 440 samples were then used for assessing the quality of the point cloud and for offline localization accuracy experiments. 

\paragraphLabel{Baseline datasets}
As RadarHD and RadCloud both used unique {\radar} configurations, we recorded additional datasets containing 10,360 (for RadarHD) and 11,244~(for~RadCloud) time-synchronized frames following similar trajectories and featuring similar environmental factors as the {\ProjectName} dataset. For the Naive {\radar} dataset, we used the existing 57,641 samples recorded for evaluating {\ProjectName}'s performance. 

\subsection{Offline Evaluations}

\paragraphLabel{Point cloud quality}
To evaluate the quality of the generated point clouds versus ground truth {\lidar} scans, we use the commonly used Chamfer and Hausdorff metrics \cite{bell_chamfer_2023,watkins_chamfer_2023,dubuisson_modified_1994}. However, given that the point clouds generated by each baseline method have varying resolutions and densities, we use the one-way version of these metrics to see how close the {\radar} point cloud is to the ground truth {\lidar} point cloud. Thus, we define the Chamfer distance (CD) as \begin{equation}
\label{eq:chamfer_distance}
    \begin{split}
       \textrm{CD}(S_{\textrm{\radar}},S_{\textrm{{\lidar}}}) = &
            \frac{1}{2|S_{\textrm{\radar}}|}\underset{{x\in S_{\textrm{\radar}}}}{\Sigma} \underset{y \in S_{\textrm{\lidar}}}{\textrm{min}}d(x,y)
    \end{split}        
\end{equation}
and Hausdorff distance (HD) as
\begin{equation*}
\label{eq:mod_haus_distance}
    % \begin{split}
        \textrm{HD}(S_{\textrm{\radar}},S_{\textrm{\lidar}})  =\underset{{x\in S_{\textrm{\radar}}}}{\textrm{max}} \underset{y \in S_{\textrm{\lidar}}}{\textrm{min}}d(x,y),
    % \end{split}
\end{equation*}
where $d(x,y)$ denotes the Euclidean distance i.e., $||x -   y||_2^2$. Finally, we record the number of points generated by each method to better understand the density of each point cloud. 

\paragraphLabel{Localization accuracy}
Next, we evaluated how well the generated point clouds could be used for downstream localization tasks. To accomplish this, we implemented a {\radar}-inertial localization stack which used an extended kalman filter (EKF) to continuously estimate the global pose of the UGV in a pre-mapped environment \cite{Kalman_1960_New}. Here, we define the pose estimate as $\poseEst{} = [\poseTransEst{},\poseHeadEst{}]$, where $\poseTransEst{} = (x,y)$ and $\poseHeadEst{} = \yaw$ denote the UGV's global position and heading, respectively. For the EKF `predict' step, we employ a non-linear first-order motion model to fuse vehicle velocity and IMU measurements. Then we used the popular iterative closest point (ICP) scan matching to ``update`` the kalman filter with new measurements. Additionally, erroneous measurements were excluded using the $\chi^2$ anomaly detector (e.g.,~\cite{Hallyburton_2022_Optimal}), with an empirically determined probability of valid data of 0.95.

To construct maps of each test environment, we utilized the popular hector mapping~\cite{kohlbrecher2011flexible} algorithm to generate a 2D map using the \lidar\ sensor. Additionally, we measured the ground truth pose $\textbf{P}$ for each experiment using a \lidar-inertial localization stack which utilized an EKF to fuse IMU, wheel encoders, and \lidar\ ICP measurements. This was benchmarked against a VICON motion capture system showing that the \lidar\ ground truth was always within \cm{20} of the VICON ground truth. 

Finally, we use the standard absolute trajectory error (ATE) and relative trajectory error (RTE) metrics~\cite{zhang2018tutorial} to assess the localization performance of each method. Here, for the $i$-th frame, the error metrics for translation (tr) and heading (hd) trajectory errors are defined as 
\begin{equation}
\label{eq:trans_traj_errors}
\begin{split}
    \textrm{ATE}_{\textrm{tr}}(i) = & ||\poseTransEst{i} - \poseTransGT{i}||_{2} \\
    \textrm{RTE}_{\textrm{tr}}(i) = &
        ||(\poseTransEst{i} - \poseTransEst{i-1}) - 
        (\poseTransGT{i} - \poseTransGT{i-1})||_{2},
\end{split}
\end{equation}
\vspace{-3.2pt}
\begin{equation}
\label{eq:head_traj_errors}
\begin{split}
    \textrm{ATE}_{\textrm{hd}}(i) = & |\poseHeadEst{i} - \poseHeadGT{i}| \\
    \textrm{RTE}_{\textrm{hd}}(i) = &
        |(\poseHeadEst{i} - \poseHeadEst{i-1}) - 
        (\poseHeadGT{i} - \poseHeadGT{i-1})|.
\end{split}
\end{equation}

\subsection{Real-time Full Stack Case Studies}
To round out our evaluation, we demonstrated the real-time feasibility of the {\ProjectName} framework, by performing real-time simultaneous localization and mapping (SLAM) and navigation case studies. We highlight that the {\ProjectName} framework was run alongside industry standard SLAM, localization, and navigation stacks in real-time on the Raspberry Pi 5. 

\paragraphLabel{Simultaneous Localization and Mapping} We used the commonly used \softwarePackage{slam-toolbox} ROS2 package to perform SLAM using the point clouds generated by the {\ProjectName} model\cite{Macenski_slam_toolbox_2021}. Due to the lower resolution nature of the radar sensing, we set the map resolution to \cm{10}.

\paragraphLabel{Navigation} Finally, we used the popular ROS2 \softwarePackage{Nav2} package to demonstrate real-time localization and navigation through a mapped environment\cite{macenski_nav2_2023survey,macenski_nav2_2020marathon2}. Here, we adapted the standard \softwarePackage{Nav2} adaptive monte carlo localization (AMCL) and default navigation configuration to utilize the enhanced point clouds generated from the {\ProjectName} framework. Noteably, we also used the maps generated in the previous step to successfully demonstrate how our framework can be used to successfully map and then autonomously navigation through an environment.

%% file: figures/tex_files/platform_overview.tex
\begin{figure}[t!]
\centering
\includegraphics[width=0.95\columnwidth]{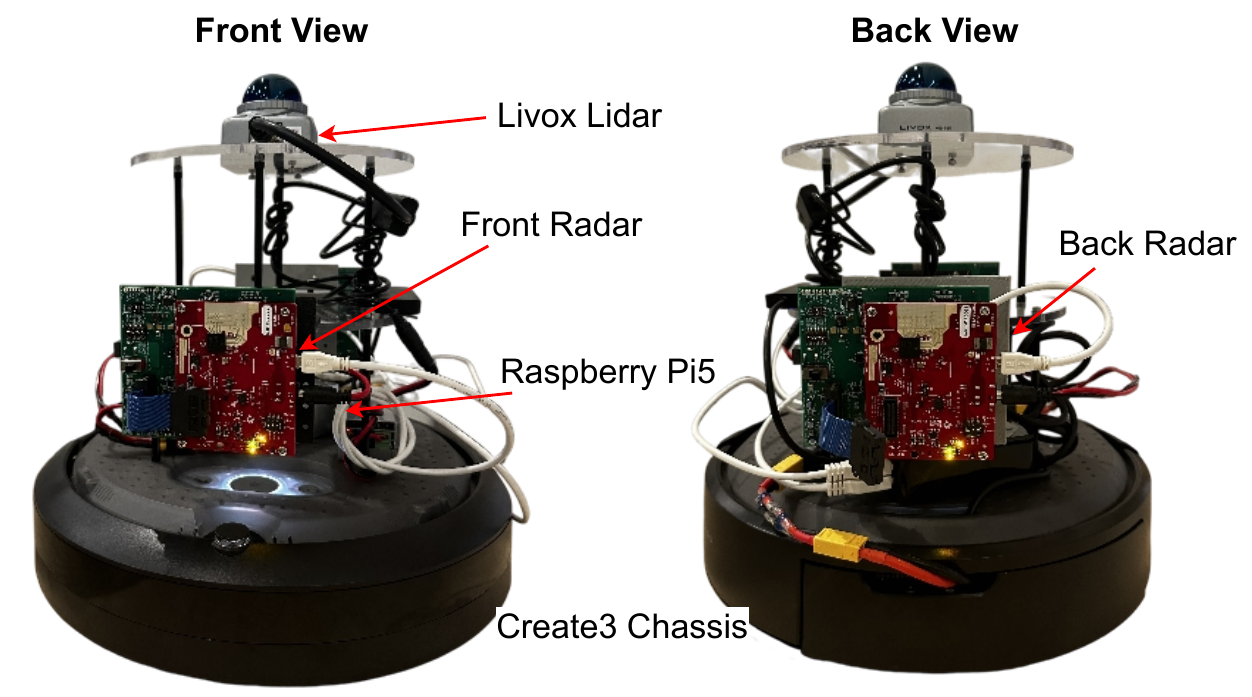}
\vspace{-4pt}
\caption{Experimental platform.}
\label{fig:platform_overview}
\vspace{-4pt}
\end{figure}

%% file: figures/tex_files/test_environments.tex
\begin{figure}[t!]
\centering
\includegraphics[width=0.95\columnwidth]{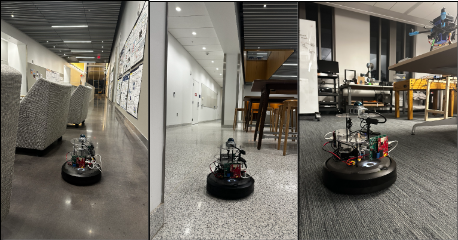}
\caption{Indoor test environments 1, 2, and 3 (left to right).}
\label{fig:test_environments}
\end{figure}

%% file: 05_results.tex
\section{Results}
\label{sec:results}
We start with an analysis of the computational time required to run each model, followed by the offline evaluations of point cloud quality and localization accuracy. Then, we conclude with a discussion on our real-time full stack case studies for SLAM and navigation. 

\subsection{Computational Time Analysis}
\input{tables/compute_time_comparison}
To compare the computational time required to make each inference on {\ProjectName} versus the other baseline methods, we conducted run-time timing tests on a Raspberry Pi 5 and a Lenovo P360 equipped with a Nvidia T1000 GPU and an Intel i9 CPU. As seen in \tbl{\ref{table:compute_time_comparison}}, {\ProjectName} significantly reduces the computational time required to enhance input point clouds. Even on a GPU equipped machine, we achieve an over 6$\times$ and 25$\times$ reduction in inference time compared to RadCloud and RadarHD, respectively. Additionally, we highlight that {\ProjectName} still maintains low inference times on the compute constrained Raspberry Pi 5, enabling real-time operations for our UGV platform.

\subsection{Offline analysis}
\input{figures/tex_files/point_cloud_comparison}
\input{tables/point_cloud_comparison}
\paragraphLabel{Point cloud quality.}
\tbl{\ref{table:point_cloud_comparison}} summarizes the point cloud quality assessment, and \fig{\ref{fig:point_cloud_comparison}} presents examples of point clouds generated using each method. As seen in \fig{\ref{fig:point_cloud_comparison}}, {\ProjectName} produces accurate and dense point clouds that feature minimal false detections. Compared to RadarHD and the Naive {\radar} methods, {\ProjectName} significantly improved the chamfer and hausdorf point cloud error metrics, indicating that its point clouds more accurately resembled the environment. Additionally, compared to the RadCloud model, {\ProjectName} produced 1.6$\times$ more points on average while still generating sufficiently accurate point clouds. Finally, we highlight that the {\ProjectName} model generalized well as it showed almost no performance drop when operating in new environments. 

\input{figures/tex_files/trajectory_comparison}
\input{figures/tex_files/ATE_error_histogram}
\paragraphLabel{Localization.}
\input{tables/localization_error_comparison}
\tbl{\ref{table:localization_traj_comparison}} summarizes the average ATE and RTE for each sensing method.  \fig{\ref{fig:trajectory_comparison}} presents an example of each method's localization performance in environment~1. As shown in \tbl{\ref{table:localization_traj_comparison}}, \ProjectName\ enables accurate location estimates of the UGV's location with an average ATE of \cm{19}. Moreover, 90\% of {\ProjectName} errors are less than \cm{35} (\fig{\ref{fig:ate_error_histogram}}). These results indicate that {\ProjectName} maintains consistently accurate trajectory estimates, even in dynamic and complex environments. 

Compared to the baseline methods, when considering the average ATE,
{\ProjectName} achieves 1.6$\times$ improvement compared to RadCloud (i.e., the best performing SOTA DL-based method), and 12.8$\times$ improvement compared to the naive \radar\ sensing method. Moreover, the naive radar baseline enabled successful localization in  most trials, but there were several trials where a large number of false detections led to a complete loss of localization; thus, resulting in the very high average ATE error.  Finally, the SOTA DL models somewhat provided accurate localization (especially RadCloud), but with sporadic large localization errors due to inaccurate predictions from the models in more complicated environments. Ultimately, {\ProjectName} provided the most accurate and consistent localization of all the considered methods.

\subsection{Real-time Full Stack Case Studies}
We now present results from our real-time SLAM and navigation case studies. Note video examples of the case studies can be found in our accompanying video submission.
\input{figures/tex_files/slam_comparison}
\input{figures/tex_files/navigation_test}
\paragraphLabel{SLAM.} \fig{\ref{fig:slam_comparison}} presents an example of a map generated using the {\ProjectName} framework. As seen in the figure, the generated map accurately captures the key features of the environment and largely resembles the map obtained using the {\lidar} sensors. Additionally, the generated map features relatively few false detections, demonstrating how {\ProjectName} enables real-time SLAM, even on compute constrained UGVs.

\paragraphLabel{Navigation.} 
Finally, we demonstrate how the maps generated using {\ProjectName}'s enhanced point clouds can be used to enable real-time {\radar} navigation. Here, \fig{\ref{fig:navigation_test}} presents an example of a navigation case study, the map of the environment, and the real-time planned trajectory of the UGV. As seen in our corresponding video submission, the UGV is successfully able to estimate its location and autonomously navigate through a complex environment by utilizing the enhanced point cloud from {\ProjectName}.

%% file: tables/compute_time_comparison.tex
\begin{table}[!t]
\begin{center}
    \caption{Comparison of average inference time where {\ProjectName} achieves significant reductions compared to existing works.}
    \label{table:compute_time_comparison}
    \begin{tabular}{ 
    c | c c c}
    \toprule
    Platform & \ProjectName & RadCloud \cite{hunt_2024_RadCloud}  & RadarHD \cite{prabhakara_high_2023}\\
    \midrule
    Raspberry Pi 5 & \msec{7.3} & \msec{178.5} & \msec{290.7} \\
    Desktop (GPU) & \msec{1.3} & \msec{8.2} & \msec{33.3} \\
    \bottomrule
    \end{tabular}
\end{center}
\end{table}

%% file: figures/tex_files/point_cloud_comparison.tex
\begin{figure}[t!]
\centering
\includegraphics[width=0.98\columnwidth]{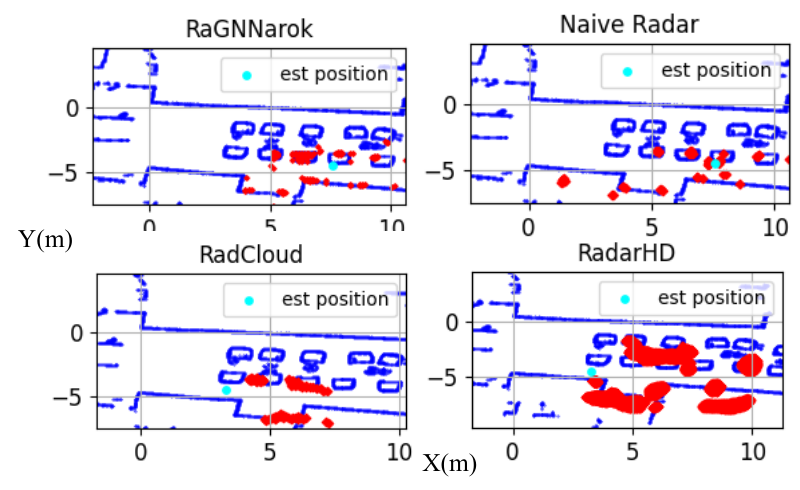}
\vspace{-8pt}
\caption{{\ProjectName} generates more accurate point clouds that remove false detections and generalize well to new environments.}
\label{fig:point_cloud_comparison}
\vspace{-4pt}
\end{figure}

%% file: tables/point_cloud_comparison.tex
\begin{table}[!t]
\begin{center}
    \caption{Comparison of point cloud quality (* and $\dagger$ indicate results from previously seen and new environments, respectively).}
    \label{table:point_cloud_comparison}
    \begin{tabular}{ 
    c | c | c c | c c}
    \toprule
    & avg & \multicolumn{2}{c|}{Chamfer (CD)} & \multicolumn{2}{c}{Hausdorf (HD)} \\
    Method & points & mean & tail (90\%) & mean & tail (90\%) \\
    \midrule
    RadarHD \cite{prabhakara_high_2023} & 5,651 & \m{0.71} & \m{1.17} & \m{3.61} & \m{6.31} \\
    RadCloud \cite{hunt_2024_RadCloud} & 65 & \m{0.25} & \m{0.42} & \m{0.85} & \m{1.54} \\
    Naive {\radar} & 65 & \m{0.47} & \m{0.71} & \m{2.30} & \m{4.02} \\
    \midrule
    {\ProjectName}* & 109 & \m{0.28} & \m{0.41} & \m{1.41} & \m{2.13} \\
    {\ProjectName}~$\dagger$ & 108 & \m{0.30} & \m{0.44} & \m{1.11} & \m{2.40} \\
    \bottomrule
    \end{tabular}
\end{center}
\end{table}

%% file: figures/tex_files/trajectory_comparison.tex
\begin{figure}[t!]
\centering
\includegraphics[width=0.98\columnwidth]{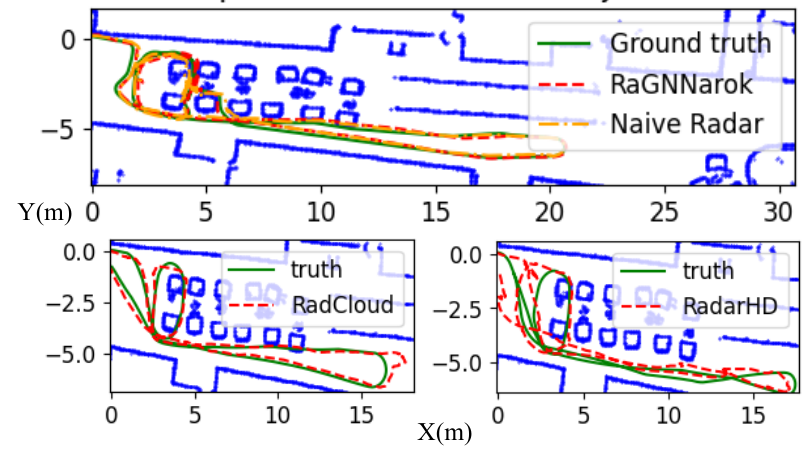}
\vspace{-8pt}
\caption{Comparison of \ProjectName and Naive Radar trajectory estimates (top), RadCloud estimates (bottom left), and RadarHD estimates (bottom right).}
\label{fig:trajectory_comparison}
\end{figure}

%% file: figures/tex_files/ATE_error_histogram.tex
\begin{figure}[t!]
\centering
\vspace{-4pt}
\includegraphics[width=0.90\columnwidth]{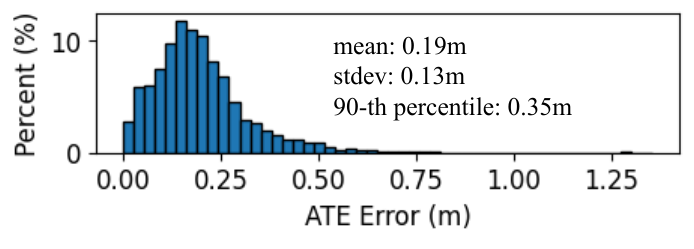}
\vspace{-8pt}
\caption{Histogram of the {\ProjectName} ATE errors illustrates how {\ProjectName} enables accurate localization performance with minimal errors}
\label{fig:ate_error_histogram}
\vspace{-4pt}
\end{figure}

%% file: tables/localization_error_comparison.tex
\begin{table}[!t]
\begin{center}
    \caption{Comparison of average trajectory errors. (* and $\dagger$ indicate results from previously seen and new environments, respectively)}
    \vspace{-8pt}
    \label{table:localization_traj_comparison}
    \begin{tabular}{ 
    c | c c | c c}
    \toprule
    & \multicolumn{2}{c|}{Absolute (ATE)} & \multicolumn{2}{c}{Relative (RTE)} \\
    Localization Method & Trans(m) & Rot(deg) & Trans(m) & Rot(deg) \\
    \midrule
    Naive Radar & \m{2.45} & \degrees{4.22} & \m{0.013} & \degrees{0.056} \\
    RadCloud \cite{hunt_2024_RadCloud} & \m{0.32} & \degrees{2.71} & \m{0.016} & \degrees{0.080} \\
    RadarHD \cite{prabhakara_high_2023}& \m{1.31} & \degrees{6.21} & \m{0.027} & \degrees{0.111} \\
    \midrule
    {\ProjectName}* & \textbf{\m{0.16}} & \textbf{\degrees{1.43}} & \textbf{\m{0.004}} & \textbf{\degrees{0.038}} \\
    {\ProjectName}$\dagger$ & \m{0.20} & \degrees{2.07} & \m{0.004} & \degrees{0.036} \\
    \bottomrule
    \end{tabular}
\end{center}
\end{table}

%% file: figures/tex_files/slam_comparison.tex
% column wide figure
\begin{figure}[t!]
\centering
\includegraphics[width=0.65\columnwidth]{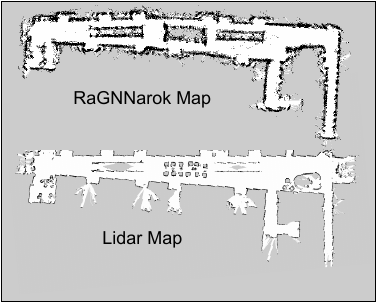}
\caption{Comparison of map generated using {\ProjectName} point clouds versus {\lidar} point clouds.}
\label{fig:slam_comparison}
\end{figure}

%% file: figures/tex_files/navigation_test.tex
% column wide figure
\begin{figure}[t!]
\centering
\includegraphics[width=0.99\columnwidth]{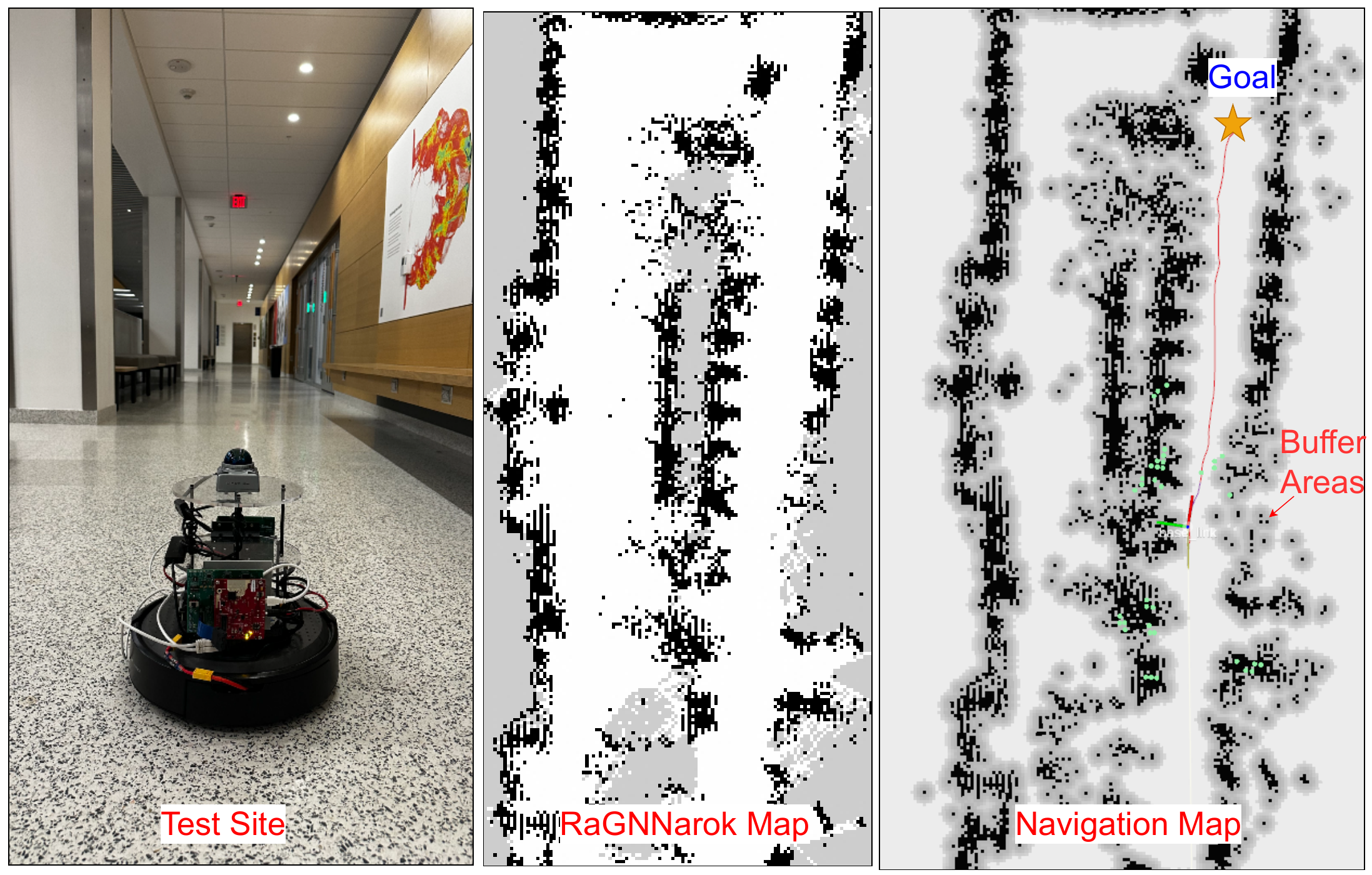}
\caption{Navigation trials in test environment 2. The green dots in the right figure denote \ProjectName-generated radar point clouds, while the red line denotes the planned trajectory.}
\label{fig:navigation_test}
\end{figure}

%% file: 06_conclusion.tex
\section{Conclusion}
\label{sec:conclusion}
% In this work, we introduced {\ProjectName} a light-weight graph neural network framework for enhancing {\radar} point clouds. Our solution is evaluated in terms of computation time per inference and point cloud quality, and we demonstrate how the enhanced point clouds can be used to enable accurate localization, mapping, and collision free navigation in complex and dynamic environments. Furthermore, we demonstrated the real-world feasibility using a real-time ROS2 unmanned ground vehicle equipped with a Raspberry Pi 5 single board computer. In future work, we will extend {\ProjectName} from 2D to 3D, as well as moving to aerial vehicles. 

In this work, we introduced {\ProjectName}, a lightweight graph neural network designed to enhance radar point clouds for real-time sensing, localization, and navigation. We evaluated its effectiveness in terms of computational efficiency and point cloud quality, as well as its ability to enable accurate localization, mapping, and collision-free navigation in complex and dynamic environments. Furthermore, we validated its real-world feasibility by deploying \ProjectName on a ROS2-based unmanned ground vehicle (UGV) running on a Raspberry Pi 5, proving that high-performance radar-based autonomy is achievable on resource-constrained platforms.Looking ahead, we plan to extend {\ProjectName} from 2D to 3D point cloud enhancement, further improving spatial awareness and precision. Additionally, we aim to adapt our framework for aerial vehicles, expanding its applicability to a broader range of autonomous systems.